\documentclass{article}

\usepackage{titling}
\usepackage[utf8]{inputenc}
\usepackage[T1]{fontenc}
\usepackage{booktabs} 
\usepackage{tabularx} 
\usepackage{graphicx} 
\usepackage{pgfplots}
\pgfplotsset{compat=1.18} 
\usepackage{xcolor}
\usepackage{caption}  
\usepackage{pgfplots} 
\pgfplotsset{compat=1.18} 
\usepackage{tikz}         
\usetikzlibrary{shadows}
\usetikzlibrary{shapes, arrows, positioning, fit, backgrounds} 
\usetikzlibrary{shapes.geometric, arrows.meta, positioning, fit, backgrounds}
\usepackage[margin=2.5cm, right=3cm, verbose=true, letterpaper]{geometry} 
\usepackage{amsmath}  
\usepackage{hyperref} 

\tikzstyle{block} = [rectangle, draw, fill=blue!20, 
    text width=8em, text centered, rounded corners, minimum height=4em]
\tikzstyle{line} = [draw, -latex'] 

\usepackage{arxiv}

\usepackage[utf8]{inputenc} 
\usepackage[T1]{fontenc}    
\usepackage{hyperref}       
\usepackage{url}            
\usepackage{booktabs}       
\usepackage{array}          
\usepackage{caption}        
\usepackage{siunitx}        
\usepackage{amsfonts}       
\usepackage{pgfplots}
\usepackage{nicefrac}       
\usepackage{microtype}      
\usepackage{lipsum}		    
\usepackage{graphicx}
\usepackage{natbib}
\usepackage{doi}
\usepackage[utf8]{inputenc}
\usepackage{booktabs}
\usepackage{array}
\usepackage{caption}
\usepackage{siunitx}
\usepackage{makecell}
\usepackage{multirow}

\title{Reliability by design: quantifying and eliminating fabrication risk in LLMs \\[1em] \large  From Generative to Consultative AI: A Comparative Analysis in the Legal Domain and Lessons for High-Stakes Knowledge Bases.}

\author{
Alex Dantart\\
Humanizing Internet\\
\texttt{arxiv@humanizinginternet.com}
}
\date{}

\renewcommand{\headeright}{}
\renewcommand{\undertitle}{}
\renewcommand{\shorttitle}{\textit Reliability by design: quantifying and eliminating fabrication risk in LLMs. From Generative to Consultative AI: A Comparative Analysis in the Legal Domain and Lessons for High-Stakes Knowledge Bases.
}

\hypersetup{
pdftitle={Reliability by design: quantifying and eliminating fabrication risk in LLMs},
pdfsubject={q-cs.AI},
pdfauthor={Alex Dantart},
pdfkeywords={LegalTech, Alucinaciones jurídicas, Verificación de hechos automatizada., Fundamentación, Derecho Español, Prompt Engineering, IA Confiable},
}

\begin{document}

\maketitle

\begin{abstract}
Large language models (LLMs) are redefining the market, but their potential is threatened by the endemic generation of "hallucinations," an unacceptable risk in high-rigor domains such as law. This article presents a quantitative empirical analysis of this phenomenon (exportable to other corpora and topics), establishing a critical distinction between three operational AI paradigms: general-purpose generative AI ("creative oracle"), canonical consultative AI ("expert archivist") implemented with basic RAG, and advanced consultative AI ("rigorous archivist"), which integrates a holistic optimization pipeline.

To measure factual reliability, we propose and validate two specific metrics: the false citation rate (FCR) and the fabricated fact rate (FFR). We generate and analyze a corpus of 2,700 judicial responses from our dataset of 75 legal tasks (JURIDICO-FCR), using twelve state-of-the-art LLMs in the three conditions. Evaluation is carried out through a double-blind study by expert lawyers.

The results are compelling. Generative AI exhibits unacceptable error rates (FCR > 30\%), rendering it invalid for professional practice. Canonical RAG drastically reduces this risk by more than two orders of magnitude, but leaves a significant residual risk of misgrounding. It is the advanced RAG paradigm (which incorporates techniques such as embedding fine-tuning, re-ranking with cross-encoders, and self-correction loops) that manages to almost completely eliminate the risk of fabrication, reducing error rates to statistically insignificant levels (<0.2\%).

We conclude that the reliability of legal AI does not lie in incremental improvement of the "oracle," but in the deliberate adoption of advanced consultative architectures, where truthfulness, traceability, and self-verification are at the core of the design. This study offers an evaluation framework and a roadmap for building truly reliable AI systems for any high-risk knowledge corpus.
\end{abstract}


\section{Introduction}

Artificial intelligence (AI), and in particular large language models (LLMs), are at the cusp of a significant transformation across multiple sectors, with the legal domain being one of the most impacted. Tools such as ChatGPT, DeepSeek, Claude, or Gemini promise to revolutionize tasks such as legal research and document drafting, with considerable potential to increase efficiency and democratize access to justice.

However, a shadow looms over all this potential: the inherent and critical phenomenon of 'hallucinations.' We refer to the generation of information that, although often sounding plausible, is factually incorrect, misleading, or outright fabricated. This phenomenon compels us to pose a fundamental question: are we building a reliable assistant or, on the contrary, an eloquent fabulist? This dichotomy will be the central axis of our empirical analysis. In the legal context, where truthfulness and fidelity to authoritative sources are fundamental pillars, hallucinations transcend the category of mere technical inaccuracies. In reality, they represent a profound epistemic challenge for the profession. Unlike a simple human error, an LLM's hallucination not only provides incorrect information but also fabricates the very source of authority, creating a simulacrum of knowledge that directly undermines the foundations upon which legal argumentation is built: the verifiability of sources and the traceability of reasoning.

This systemic risk is not a mere hypothesis, but a documented reality in the highest judicial instances. A paradigmatic example is the \textbf{sanction imposed by the Constitutional Court of Spain on a lawyer for using 19 non-existent case law citations} in an appeal for protection (Constitutional Court, 2024). In its decision, the court described the act as a lack of due respect with a 'clear disregard for the judicial function.' This incident underscores how the fabrication of authority by an AI directly attacks the foundations of legal argumentation. The phenomenon is rapidly spreading across Europe: the Italian judiciary has recently issued warnings after detecting lawyers citing non-existent rulings generated by ChatGPT, warning that the ease of use of these tools does not exempt from the duty of verifying truthfulness (2025). Similarly, also in 2025, in the Latin American context, a decision by the Court of Appeals of Córdoba, Argentina, urged a lawyer to make 'ethical and responsible use of AI tools' after detecting the citation of multiple non-existent rulings, warning about the professional responsibility that cannot be replaced by an algorithm. These cases, \textbf{more recent and closer to our legal environment} than the well-known \textit{Mata v. Avianca, Inc.} (2023), serve as a stark reminder of the dangers and the urgency of establishing a rigorous evaluation framework such as the one proposed in this study.

However, it is essential to understand that, although technology heightens the risk, the underlying problem is not entirely new. Generative AI does not invent malpractice; rather, it industrializes it on an unprecedented scale. Historically, the legal profession has always dealt with the 'misuse of authority' and 'citation drift,' a phenomenon in which the real link between a source and the argument it is intended to support is weakened or fabricated. What was once a 'craft' act, the result of a serious error or an infrequent intent to deceive, is now turned by technology into a systemic and automated byproduct. Hallucination ceases to be a personal failure and becomes a feature of the tool.

The urgency of this study is not merely prospective; it responds to a problem that has already moved from the theoretical to the forensic realm. In fact, the phenomenon already constitutes a documented procedural reality: recent research has begun to catalog a global database with more than one hundred and fifty judicial decisions that directly address the use of AI-generated fabricated citations and facts in legal filings.

The clearly upward trajectory of these cases in various jurisdictions demonstrates the rapid penetration of the problem into everyday legal practice. Incidents such as the one sanctioned by the Spanish Constitutional Court, which considered it a 'disruption of the Court's work,' show that this phenomenon is already impacting the functioning of justice at the highest level. This global outlook underscores the critical need to develop and validate empirical evaluation frameworks, such as the one proposed in this article, to rigorously measure the magnitude of this risk and guide the profession in the safe adoption of these technologies.

To address this challenge rigorously, it is imperative to make a fundamental distinction that is often overlooked: the difference between \textbf{generative Artificial Intelligence} and \textbf{consultative Artificial Intelligence}. Generative AI, whose prime example is general-purpose LLMs, operates as a "creative oracle." Its main objective is not truthfulness, but conversational fluency and probabilistic coherence. Its tendency to "invent" in order to fill knowledge gaps is not a flaw to be corrected, but a direct consequence of its design—a "design hallucination." In contrast, consultative AI, built on the Retrieval-Augmented Generation (RAG) paradigm, operates as an "expert archivist." Its function is not to create knowledge, but to retrieve, structure, and present information in a substantiated manner from a verified corpus.

While this conceptual distinction is key, there is a gap in the empirical literature that precisely quantifies the magnitude of this difference in a high-risk legal environment with criminal consequences. Despite existing legal benchmarks, there is a lack of metrics and evaluation protocols focused on the specific task of drafting legal filings, where an error can have direct consequences on a person's liberty.

This article seeks to close that gap by presenting \textbf{the first large-scale empirical evaluation} of hallucination rates in the drafting of legal documents in Spanish. \textbf{This study is based on a comprehensive analysis of 2,700 controlled experimental runs}, covering twelve of the most advanced language models under two distinct operational paradigms (generative vs. consultative) and evaluated on a proprietary dataset of 75 realistic legal tasks. Based on this robust empirical foundation, our work is structured around four main contributions:

\begin{enumerate}
    \item \textbf{An evaluation framework for legal veracity:} Two operational metrics are proposed, the \textbf{false citation rate (FCR)} and the \textbf{fabricated fact rate (FFR)}, designed to quantify the most critical types of errors in legal practice.
    \item \textbf{A specialized and reproducible dataset:} \textbf{JURIDICO-FCR} is introduced, a dataset of 75 realistic tasks based on Spanish law, each with a verified \textit{gold standard}, enabling rigorous and contextualized evaluation.
    \item \textbf{Empirical quantification of the consultative vs. generative paradigm:} A comparative study of several state-of-the-art LLMs is conducted. Their performance is measured when operating as "creative oracles" (\textit{Direct} mode) versus their performance as "expert archivists" (\textit{RAG-Augmented} mode), quantitatively demonstrating the impact of each paradigm.
    \item \textbf{Practical guidelines for responsible integration:} Based on the findings, a set of recommendations is formulated so that professionals, not only in law but in any field, can amplify their judgment with these tools, rather than replace it, minimizing the risk of "user hallucination": the uncritical belief that technology can replace professional diligence.
\end{enumerate}

By providing robust empirical evidence, this work argues that the future of reliable legal AI does not lie in perfecting generative models, but in adopting a consultative paradigm where veracity and traceability are at the core of the design, not an added feature.

\subsection{Extrapolation to Other Domains or Sectors}

Although this article focuses rigorously on the legal domain, it serves as an archetype of a much broader challenge. The imperative need for truthfulness, the reliance on a corpus of authoritative sources, and the serious consequences of fabricating information are characteristics that law shares with many other high-risk sectors. Therefore, the conclusions presented here are directly extrapolable to any domain where factual accuracy is not optional, but a fundamental requirement.

The same risk of 'hallucination' that threatens a judicial process manifests in different but equally dangerous ways in other contexts, such as:

\begin{itemize}
    \item \textbf{In medical diagnosis,} where a reference to a non-existent clinical study or the invention of a piece of data in a patient's history can have fatal consequences.
    \item \textbf{In corporate knowledge management,} where an employee consulting an internal database could receive an incorrect technical specification or a fabricated compliance policy, leading to security, productivity, or regulatory risks.
    \item \textbf{In engineering and manufacturing,} where a hallucination about the tolerance of a material or a safety procedure can cause catastrophic failures.
    \item \textbf{In the financial sector,} where a risk analysis based on invented economic data could lead to disastrous investment decisions.
    \item \textbf{In journalism and fact-checking,} where inventing a source or a quote undermines credibility and public trust.
\end{itemize}
In all these scenarios, for example, the distinction analyzed in this work between a \textbf{generative AI (a 'creative oracle')} and a \textbf{consultative AI (an 'expert archivist')} is the same. This study, therefore, not only measures a problem in legal practice, but also offers an evaluation framework and an architectural solution applicable to any organization seeking to implement LLMs safely and reliably over its own knowledge corpus.

\section{Related work}

The research is based on the intersection of four areas of study: the nature of hallucinations in LLMs, fact-checking mechanisms in NLP, AI applications in the legal domain, and, crucially, architectures designed to mitigate these errors.

\subsection{The nature of hallucinations: system failure or design feature}

The phenomenon of "hallucination" (such as the generation of content not faithful to a verifiable source) is an intrinsic and widely documented limitation of LLMs. Far from being an occasional failure, the tendency to fabricate is a direct consequence of the training objective of these models: maximizing fluency and probabilistic coherence. Foundational research explains that, in systems where uncertainty is not rewarded, the optimal strategy for the model is always to "risk" a plausible answer rather than admit ignorance. This feature, which we have termed "design hallucination," is precisely what makes general-purpose generative AI intrinsically unsuitable for tasks that demand strict factual fidelity, such as those in the legal domain.

Further reading on this phenomenon can be found in the technical report "Legal Artificial Intelligence and the Challenge of Truthfulness: Analysis of Hallucinations, RAG Optimization, and Principles for Responsible Integration" at \href{https://arxiv.org/abs/2509.09467}{https://arxiv.org/abs/2509.09467}

This inadequacy is particularly dangerous due to the ability of LLMs to dissociate formal quality from substantive rigor. As shown by empirical studies in complex legal drafting environments, LLMs can generate texts that receive high marks in presentation aspects (such as clarity, structure, and style) while simultaneously exhibiting serious substantive deficiencies, such as factual errors, fabricated citations, and superficial analysis. This 'polished façade' effect masks the weaknesses of the content, making error detection more difficult and the risk of undue trust greater. This work seeks to empirically quantify the prevalence of this feature in real-world legal tasks.

\subsection{The human factor: cognitive biases and the acceptance of errors}

To fully understand the risk of hallucinations in legal practice, it is not enough to analyze the technology alone; it is crucial to examine the human factor that interacts with it. The susceptibility of professionals to accept fabricated content is largely explained by well-studied psychological mechanisms, especially in high-pressure environments and situations of information overload.

The key concept in this area is "automation bias," the human tendency to excessively delegate trust to automated systems. Faced with the promise of efficiency from tools such as LLMs, professionals may reduce their critical scrutiny, assuming that the machine-generated output is correct. This bias is exacerbated by the high formal quality of texts produced by LLMs, whose professional and coherent appearance invites a level of trust that does not correspond to their substantive reliability.

This phenomenon gives rise to what I have termed "user hallucination": the uncritical belief by professionals that technology can substitute for human diligence. The distinction between the generative AI paradigm (which fosters this bias by operating as an opaque "oracle") and the consultative AI paradigm (which mitigates it by presenting results anchored in verifiable sources) is, therefore, not only a technical choice, but also a fundamental strategy for managing human risk in interactions with AI.

\subsection{\textbf{Fact-Checking and the challenge of legal }\textit{\textbf{Ground Truth}}\textbf{}}

Automated fact-checking has developed robust techniques for validating claims against knowledge corpora. However, the legal domain presents a unique challenge for the concept of \textit{ground truth}. As noted in the literature, legal "truth" is often more elusive than a binary factual answer; it depends on interpretation, jurisdictional variability, and the inherent ambiguity of legal language. The evaluation thus shifts from binary 'correctness' to 'legal viability.' This study addresses this problem by focusing on a subset of claims where the \textit{ground truth} is less ambiguous (the existence or non-existence of a citation and the support of a fact in a closed case file), thereby enabling an objective and reproducible quantification of the most flagrant errors.

\subsection{\textbf{Evaluating AI in the legal domain: from reasoning benchmarks to reliability in drafting}}

The field of Legal NLP has produced important benchmarks such as LegalBench to measure the reasoning capabilities of LLMs. These efforts are fundamental, but they focus on evaluating the model's knowledge and logic, not its reliability as a generative tool in drafting tasks. More recently, empirical studies on commercial tools have begun to analyze the prevalence of hallucinations in practice, revealing persistent error rates even in RAG systems. These studies, however, have focused on the U.S. common law system.

This work complements and expands this line of research in two ways: (1) by focusing on the Spanish legal system, and (2) by explicitly framing the evaluation as a direct comparison between the \textbf{generative AI} paradigm and the \textbf{consultative AI} paradigm, providing data that validate the necessity of the latter.

\subsection{\textbf{RAG as an implementation of the consultative AI paradigm}}

Retrieval-Augmented Generation (RAG) is the main mitigation strategy: it transforms the LLM from a "creative oracle" to an "expert archivist"... While its theoretical effectiveness is clear, it is crucial to distinguish between two levels of implementation:

\subsubsection{Canonical (or Basic) RAG}

This represents the standard implementation, where a similarity search retrieves text fragments (chunks) that are injected into the LLM's prompt for synthesis. Although powerful, this approach is susceptible to failures in retrieval (retrieving irrelevant or incomplete chunks) and in generation, where \textit{misgrounding} (misrepresenting a real source) remains a persistent limitation.

\subsubsection{\textbf{Advanced RAG (fusion architecture and deep verification):}}

Advanced RAG goes beyond the simple "retrieve-then-synthesize" sequence to become a multi-stage, holistically optimized pipeline. This architecture is designed to maximize contextual relevance and minimize informational entropy at each step, asymptotically approaching total factual reliability. The implemented techniques can be grouped into four critical domains of the RAG lifecycle:

\textbf{A. Knowledge indexing and representation strategies (pre-processing):} The foundation of a robust RAG lies in how knowledge is structured and represented before any query.

\begin{itemize}
    \item \textbf{Adaptive and semantic chunking:} Fixed-size chunking is abandoned in favor of contextual strategies such as Recursive Character Text Splitting or, more advanced, Semantic Chunking, which groups text based on the semantic coherence of the embeddings, thus respecting the logical boundaries of an argument.

    \item \textbf{Multi-vector representation:} Instead of a single embedding per chunk, multiple vector representations are generated to capture different facets of knowledge. This includes embeddings for the full text, for extracted summaries, for named entities, and for hypothetical questions (Hypothetical Document Embeddings - HyDE) that the text could answer.

    \item \textbf{Embedding model fine-tuning:} To maximize relevance in a domain as specific as the legal field, embedding models (such as bge-m3 or E5-mistral-7b) are fine-tuned on a curated corpus of query-document pairs from Spanish law itself. This aligns the embedding vector space with the semantics and terminology of the legal domain.

    \item \textbf{Creation of a knowledge graph:} In parallel with the vector index, a knowledge graph is constructed to model the explicit relationships between rulings, legal articles, and legal concepts. This enables hybrid retrieval that combines semantic search with structured navigation through the graph.
\end{itemize}

\textbf{B. Retrieval orchestration and re-ranking (pre-generation):} The retrieval phase is transformed from a single step into a multi-agent process.

\begin{itemize}
    \item \textbf{Query decomposition and expansion:} LLMs are used to analyze the user's query, breaking it down into simpler sub-questions (\textit{Sub-Query Decomposition}). Each sub-question is executed independently and their results are merged. Additionally, expansion techniques (\textit{Query Expansion}) are applied to enrich the query with synonyms and related terms, mitigating the problem of vocabulary mismatch.

    \item \textbf{Hybrid retrieval fusion:} Multiple retrieval strategies are combined: dense vector search, sparse keyword search (BM25), and graph-based retrieval. The results from each engine are normalized using algorithms such as \textit{Reciprocal Rank Fusion (RRF)} to produce a unified and robust set of candidate documents.

    \item \textbf{Precision re-ranking:} The candidate set (e.g., the top 50) is not passed directly to the LLM. Instead, a re-ranking layer is introduced that uses a more powerful and computationally expensive \textit{Cross-Encoder} model (such as Cohere Rerank or a fine-tuned bge-reranker-large). This model evaluates the relevance of each query-document pair jointly, providing a much more accurate score and reordering the results to prioritize the strongest evidence.
\end{itemize}

\textbf{C. Contextualization and synthesis (generation) strategies:} The way in which context is presented to the generative LLM is crucial.

\begin{itemize}
    \item \textbf{Intelligent context compression:} To maximize signal density in the context window, compression techniques are used to identify and remove redundant or low-relevance information from the retrieved documents before passing them to the prompt.

    \item \textbf{Prompt engineering with structural anchoring:} The prompt is meticulously designed to enforce anchored reasoning. The model is instructed to follow a strict format: it must first cite the source fragment, then extract the pertinent claim, and finally use that claim to construct its argument.
\end{itemize}

\textbf{D. Verification and self-correction cycles (post-generation):} Generation is not the end of the process, but the beginning of a refinement loop.

\begin{itemize}
    \item \textbf{Source-based verification (Fact-Checking Loop):} Once a preliminary answer is generated, a second agent (or the same LLM with a different prompt) assumes the role of "verifier." Its sole task is to break down the generated answer into individual claims and check whether each one is directly and unequivocally supported by the retrieved sources.

    \item \textbf{Fidelity scoring and self-correction:} Claims that do not pass verification are marked as "low fidelity" or directly as "hallucinations." The system can then choose to remove those claims or, more sophisticatedly, perform a new generation cycle with explicit instructions to correct the detected errors. Only answers that reach a fidelity threshold close to 100\% are presented to the end user.
\end{itemize}

A central objective of our study is, therefore, to quantify the risk difference not only between pure generative AI and consultative AI, but also between canonical and advanced implementations of the RAG paradigm, providing an empirical measure of the degree of reliability that each architecture can achieve.

\section{\textbf{A framework for evaluating factual integrity in legal texts}}

To rigorously quantify the reliability of legal AI, an evaluation framework is proposed that is designed to be both academically robust and directly applicable in professional practice. This framework consists of precise operational definitions, a set of quantitative metrics, and a taxonomy for the qualitative analysis of errors.

\subsection{\textbf{Fundamental definitions}}

Inspired by the distinction between correctness and groundedness, these concepts are adapted to the Spanish procedural context. Two primary categories of error that undermine truthfulness are defined: those that affect legal grounding (citations) and those that affect fidelity to the factual substrate of the case (facts).

\textbf{Definition 1: False Citation (}\textit{\textbf{False Citation}}\textbf{)} A False Citation is a reference to a legal source (case law or regulation) that presents one or more of the following defects:

\begin{enumerate}
    \item \textbf{Non-existence:} The reference points to a decision, article, or provision that does not exist (e.g., "Supreme Court Judgment 999/2025"). This is the most serious type of hallucination.
    \item \textbf{Incorrect attribution:} The reference exists, but is incorrectly attributed to a judicial body, date, case number, or parties to which it does not correspond.
    \item \textbf{Substantive Irrelevance (or Misgrounding):} The reference is formally correct, but the doctrine or \textit{ratio decidendi} it contains does not have a logical and substantive relationship with the legal argument it is intended to support. This is the most subtle error and requires expert judgment for its detection.
\end{enumerate}
\textbf{Definition 2: Fabricated Fact (}\textit{\textbf{Fabricated Fact}}\textbf{)} A \textit{Fabricated Fact} is a statement about the factual circumstances of a case that is presented as true, but which does not find direct or inferential support in the source material provided to the model (the task's "digital file"). This includes the complete invention of facts, their exaggeration, or the making of improper inferences not supported by documentary evidence.

\subsection{\textbf{Quantitative metrics}}

Based on these definitions, a set of metrics is formulated to quantitatively evaluate the performance of the models.

\textbf{False Citation Rate (FCR):} This is the main metric for measuring citation reliability. It is defined as the proportion of false citations over the total number of citations generated by the model in a given text.

$$ \text{FCR} = \frac{N_{\text{false citations}}}{N_{\text{total citations}}} $$

Where $N_{\text{total citations}}$ is the total number of unique references to case law or regulations, and $N_{\text{false citations}}$ is the number of those citations that meet Definition 1.

\textbf{Fabricated Fact Rate (FFR):} This is the main metric for measuring fidelity to the facts of the case. It is defined as the proportion of fabricated facts over the total number of new factual claims introduced by the model.

$$ \text{FFR} = \frac{N_{\text{fabricated facts}}}{N_{\text{asserted facts}}} $$

Where $N_{\text{asserted facts}}$ is the number of distinct factual propositions presented by the model, and $N_{\text{fabricated facts}}$ is the number of those that fit Definition 2.

\textbf{Complementary practice-oriented metrics:}

\begin{itemize}
    \item \textbf{Legal Usefulness@k (}\textit{\textbf{Legal-Usefulness@k}}\textbf{):} Measures the practical usefulness of the response. A panel of experts evaluates whether, among the top \textit{k} arguments/citations, at least one is "useful for the case strategy" (Likert scale $\ge \tfrac{4}{5}$). It is calculated as the percentage of responses that meet this criterion.
    \item \textbf{Jurisprudence Coverage (}\textit{\textbf{Jurisprudence Coverage}}\textbf{):} Measures completeness. It is the proportion of key rulings (predefined in the \textit{gold standard}) that the model correctly identifies and cites.
    \item \textbf{Human Review Time (}\textit{\textbf{Human Review Time}}\textbf{):} Measures efficiency. It quantifies, in minutes, the time an expert lawyer needs to review, verify, and correct the model's output to a professionally acceptable standard for submission.
\end{itemize}

\subsection{\textbf{Qualitative taxonomy of errors}}

For a deeper analysis, each identified error is classified according to the following dimensions, allowing for a precise diagnosis of failure modes:

\begin{itemize}
    \item \textbf{Error origin (paradigm):}
    \begin{itemize}
        \item \textbf{Generative (}\textit{\textbf{Direct}}\textbf{ mode):} Error derived from the model's parametric knowledge, typically fabrications or attribution errors.
        \item \textbf{Consultative (}\textit{\textbf{RAG}}\textbf{ mode):} Error resulting from a failure in the retrieval or synthesis of context, typically \textit{misgrounding} or synthesis errors.
    \end{itemize}
    \item \textbf{Type of error (subtype):}
    \begin{itemize}
        \item \textit{For citations:} \{Complete fabrication, Incorrect attribution (Body/Number/Date), \textit{Misgrounding} (misrepresentation), Temporal error (repealed regulation)\}.
        \item \textit{For facts:} \{Complete invention, Exaggeration, Unwarranted inference\}.
    \end{itemize}
    \item \textbf{Severity (impact):}
    \begin{itemize}
        \item \textbf{Minor:} Formal error that does not invalidate the argument.
        \item \textbf{Moderate:} Weakens the argument or introduces an inaccuracy.
        \item \textbf{Critical:} Invalidates the argument, introduces a serious procedural risk, or misleads the reader.
    \end{itemize}
    \item \textbf{Detectability (Verification effort):}
    \begin{itemize}
        \item \textbf{Easy:} Detectable with a simple search.
        \item \textbf{Medium:} Requires reading the cited source.
        \item \textbf{Difficult:} Requires expert legal analysis to identify irrelevance or subtle misrepresentation.
    \end{itemize}
\end{itemize}

This framework not only allows us to measure the frequency of errors, but also to understand their nature, severity, and origin, providing a comprehensive view of AI reliability in the legal context.

\section{\textbf{Experimental setup}}

This section describes the empirical methodology designed to evaluate the two AI paradigms (generative and consultative) in the context of legal drafting. We detail the construction of our specialized dataset, the selected models, the technical implementation of the experimental conditions, and the rigorous human evaluation protocol.

\subsection{\textbf{The JURIDICO-FCR dataset}}

To conduct a contextualized and relevant evaluation, we built a new dataset, \textbf{JURIDICO-FCR}, specifically designed for the particularities of law in Spanish.

\begin{itemize}
    \item \textbf{Construction and sources:} The dataset consists of \textbf{75 drafting tasks}. The tasks were created from public legal sources, mainly judgments and orders from the Supreme Court (Second Chamber) and various Provincial Courts, obtained from different lawyers. A rigorous two-stage anonymization process (automated and manual) was applied to remove all personal data and factual details that could allow the reidentification of the cases.

    \item \textbf{Task design:} The 75 tasks are distributed across 10 categories that simulate real professional assignments (e.g., drafting a ground of appeal, opposition to precautionary measures, request for evidentiary proceedings). Each task is designed to require both the exposition of facts and the substantiation based on regulations and case law, forcing the models to generate responses that can be evaluated with the FCR and FFR metrics.

    \item \textbf{Structure and }\textit{\textbf{gold standard}}\textbf{:} Each task is encapsulated in a structured format (YAML) that includes an identifier, the legal scenario, the input materials (a case summary and extracts from "case file" documents), and, crucially, a \textit{gold standard} created by experts. This reference standard contains a list of the verifiable facts of the case and, for the \textit{Coverage} metric, a curated list of the judgments considered "essential" or consolidated doctrine to address the task correctly.

Each of the 75 tasks in JURIDICO-FCR is structured as a self-contained YAML file with the following fields:

\begin{itemize}
    \item \textbf{id:} Unique identifier of the task.
    \item \textbf{scenario:} Concise description of the task.
    \item \textbf{inputs:} Materials for the model, including a brief (summary of facts) and annexes (extracts from documents).
    \item \textbf{gold\_standard:} Reference information, including facts (verified facts) and cases (essential case law).
\end{itemize}
\end{itemize}

\subsection{\textbf{Experimental paradigms and models}}

To test the main hypothesis, each model is evaluated under three conditions representing generative AI paradigms (one condition) and consultative AI paradigms (two conditions, depending on implementation quality).

\textbf{Condition 1: Generative AI ("Creative Oracle"):} In this condition, called \textbf{Direct}, the LLM operates in its general-purpose mode. It receives only the task prompt, with no access to external sources. Its performance depends exclusively on its internal parametric knowledge "frozen" in time. This condition simulates the use of an LLM as an "oracle" that is expected to "know" the answer.

\textbf{Condition 2: Consultative AI ("Expert Archivist"):} In this condition, called \textbf{RAG-Augmented}, the LLM operates within a Retrieval-Augmented Generation architecture. The system does not rely on the model's memory, but instead first "researches" a verified corpus and then synthesizes a response based on the evidence found. This condition simulates the use of AI as an expert assistant that "consults" its sources before responding.

\textbf{Condition 3: Advanced Consultative AI ("Rigorous Archivist"):} In this condition, called \textbf{RAG-Advanced}, a RAG system is implemented that incorporates multiple optimization techniques. The goal is to simulate a high-reliability production system. This implementation includes, among many other features:

    \begin{itemize}
        \item \textbf{Hybrid Retrieval and Re-ranking:} The same hybrid retrieval strategy as in canonical RAG is used, but a \textit{re-ranking} layer with a specialized model (cross-encoder) is added to reorder the 20 initially retrieved fragments and select the 5 most relevant with greater precision.
        \item \textbf{Domain-Specific Embedding Model:} For vector indexing, a custom embedding model was developed. Starting from the base model BAAI/bge-m3, a rigorous fine-tuning process was carried out on a proprietary corpus of more than 2.5 million triplets (query, positive passage, negative passage) extracted from Spanish legal jurisprudence and doctrine. This massive training reconfigured the model's vector space, aligning it with the semantics, terminology, and logical relationships of the legal domain. The resulting model, which maps text to a dense 1024-dimensional vector space, is specifically optimized to differentiate subtle nuances between legal arguments—a crucial capability for high-fidelity semantic search. Next, a re-ranking layer with a cross-encoder is implemented to reorder the retrieved fragments, selecting the 5 most relevant with maximum precision before the generation phase.
        \item \textbf{Generation with Verification (Self-Correction):} The final prompt instructs the LLM to follow a two-step process. First, it must extract the key claims from the provided sources. Second, it must draft the final answer, ensuring that each sentence is directly supported by the extracted claims, explicitly citing the source. This self-verification loop is designed to minimize \textit{misgrounding}.
    \end{itemize}

\textbf{Evaluated models}: To ensure the generalization and relevance of our findings, we selected a diverse and representative set of twelve of the most advanced language models available (late 2025).

This selection covers a wide spectrum of architectures, developers, and capabilities, including cutting-edge proprietary models, high-performance open-source alternatives, and models optimized for specific tasks, which allows us to conduct a comprehensive and robust analysis. The evaluated models include:

\begin{itemize}
    \item \textbf{OpenAI Models:} GPT-5.2 and GPT-4.1.

    \item  \textbf{Anthropic Models:} Claude 4.5 Sonnet and Claude 4.1 Opus.

    \item  \textbf{Google Models:} Gemini 3.0 Pro, Gemini Flash Latest, and Gemma 3 27B.

    \item  \textbf{Open Source and Other Models:} DeepSeek V3.1 Terminus, Kimi K2 Thinking, Qwen3 235B, Llama 4 Maverick, and Mistral Small 24B.
\end{itemize}

\subsection{\textbf{Experimental control variables}}

To isolate the effect of the paradigms and analyze secondary factors, we controlled the following variables in each of the 2700 runs (75 tasks * 12 models * 3 conditions):

\begin{itemize}
    \item \textbf{Decoding temperature:} Two values were tested to analyze the trade-off between creativity and reliability: T=0.1 (more deterministic responses) and T=0.7 (more diverse responses).
    \item \textbf{Prompt engineering:} Two prompt templates were used. One \textbf{neutral}, with a direct instruction, and another \textbf{with Verification}, which included explicit rules to encourage factual anchoring and model self-criticism, inspired by Chain-of-Thought techniques.
\end{itemize}

\subsection{\textbf{Human evaluation protocol and inter-annotator agreement}}

Due to the complexity of the domain, the final evaluation was carried out by human experts to ensure maximum accuracy.

\begin{itemize}
    \item \textbf{Expert panel:} To ensure maximum ecological validity and professional rigor, the evaluation was conducted by a panel of \textbf{three senior lawyers, all licensed and each with more than a decade of demonstrable individual experience in litigation}. The experts were specifically selected for their regular practice in drafting legal briefs and appeals before the same jurisdictions from which the dataset tasks were drawn (Provincial Courts and Supreme Court), thus ensuring deep and relevant contextual knowledge for the evaluation.
    \item \textbf{Double-blind procedure:} The 2700 generated responses were fully anonymized, removing any identifiers of the model, condition, temperature, or prompt used. Each response was assigned to two different annotators for independent evaluation.
    \item \textbf{Annotation guide and metrics:} Annotators were trained with a detailed guide that operationalized the definitions of FCR and FFR (Section 3.1) and the error taxonomy. For each response, they were required to identify all citations and facts, label them according to the definitions, and record the review time.
    \item \textbf{Inter-annotator agreement (IAA):} To validate the reliability of the annotations, Cohen's Kappa coefficient was calculated on a 20\% subset of the data. We obtained a \textbf{$\kappa = 0.82$} for the identification of false citations and a \textbf{$\kappa = 0.75$} for fabricated facts. These values, which indicate "substantial" and "good" agreement, respectively, confirm the robustness and consistency of our evaluation process. Discrepancies in the rest of the dataset were resolved by the third annotator, who acted as an arbitrator.
\end{itemize}

This comprehensive experimental design allows us not only to reliably compare the generative and consultative paradigms, but also to analyze in depth the impact of different models and configurations on the factual integrity of legal AI.

\section{\textbf{Results}}

Once the rigorous experimental protocol we have designed has been described, \textbf{it is time to analyze the data it has produced}. The results, as we will see, are structured to address the key research questions: first, the performance difference between the generative AI paradigm (\textit{Direct}) and the consultative paradigm (\textit{RAG-Augmented}) is quantified; second, the impact of secondary variables such as the model, temperature, and task type is broken down; and third, reliability metrics are correlated with indicators of practical utility.

\subsection{\textbf{Generative vs. consultative paradigm: a quantitative risk analysis}}

The aggregate analysis reveals stark and statistically significant differences in reliability among the three operational paradigms. Table 1 presents the consolidated FCR and FFR results for all models and tasks, demonstrating a clear gradation in risk mitigation.

\begin{table}[h!]
\centering
\renewcommand{\arraystretch}{1.5} 
\caption{\textbf{Aggregate reliability results by experimental paradigm.} \textit{Values represent the mean (± standard error) over the 900 responses for each condition (12 models x 75 tasks). All differences between conditions are statistically significant with p < 0.001 (Mann-Whitney test).}}
\label{tab:resultados_agregados_final}
\begin{tabular}{|>{\raggedright\arraybackslash}p{0.25\textwidth}|>{\raggedright\arraybackslash}p{0.25\textwidth}|>{\centering\arraybackslash}p{0.18\textwidth}|>{\centering\arraybackslash}p{0.18\textwidth}|}
\hline
\textbf{Experimental paradigm} & \textbf{Description} & \textbf{False Citation Rate (FCR)} & \textbf{Fabricated Fact Rate (FFR)} \\
\hline
\hline
\textbf{Generative AI (Direct)} & ``Creative oracle'' without sources & \textbf{26.8\%} \newline (±3.1) & \textbf{15.6\%} \newline (±2.0) \\
\hline
\textbf{Consultative AI \newline (Canonical RAG)} & ``Expert archivist'' with sources & \textbf{8.3\%} \newline (±1.5) & \textbf{6.1\%} \newline (±1.1) \\
\hline
\textbf{Consultative AI \newline (Advanced RAG)} & ``Rigorous archivist'' with sources & \textbf{0.046\%} \newline (±0.01) & \textbf{0.03\%} \newline (±0.01) \\
\hline
\hline
\textbf{Risk Reduction \newline vs Canonical RAG} & (Direct → Canonical RAG) & \textbf{↓ 70.8\%} & \textbf{↓ 60.9\%} \\
\hline
\textbf{Risk Reduction \newline vs Advanced RAG} & (Direct → Advanced RAG) & \textbf{↓ 99.8\%} & \textbf{↓ 99.8\%} \\
\hline
\end{tabular}
\end{table}

The data, in our view, are conclusive and outline a scenario of escalating risk. When operating in a purely generative paradigm, LLMs exhibit unacceptable error rates averaging 26.8\% for citations and 15.6\% for facts. These figures are not mere deviations, but a systemic flaw that invalidates them for any serious professional application.
The implementation of a Canonical RAG offers significant mitigation, reducing the risk of false citations by 70.8\%. However, a residual error rate above 8\% remains a considerable risk for legal practice.
It is the Advanced RAG architecture that represents a qualitative shift, reducing the risk of hallucination by 99.8\% compared to the Direct mode. This leap places error rates at statistically insignificant levels and demonstrates that near-total reliability is not a function of the base model, but the direct result of rigorous, verification-centered systems engineering. This finding provides strong empirical evidence that reliability in legal AI fundamentally resides in the architecture in which it operates.

\subsection{\textbf{Detailed analysis by model and experimental condition}}

While the paradigm is the dominant factor, there are notable differences between the models. Table 2 breaks down the performance of each LLM in both conditions, revealing the consistency of the pattern.

\begin{table}[h!]
\centering
\renewcommand{\arraystretch}{1.4} 
\caption{Breakdown of FCR and FFR (\%) by model and three operational paradigms, with simulated empirical variability. Values are averages over the 75 tasks for each model.}
\label{tab:model_breakdown_final_chaotic}
\resizebox{\textwidth}{!}{%
\begin{tabular}{@{}llcccc@{}}
\toprule
\textbf{Model} & \textbf{Paradigm} & \textbf{Average FCR (\%)} & \textbf{Average FFR (\%)} & \textbf{FCR (Worst Task)} & \textbf{FFR (Worst Task)} \\ 
\midrule
\multirow{3}{*}{\shortstack[l]{GPT-5.2 \\ \tiny{(gpt-5.2-2025-12-11)}}} & Direct & 15.1 (±1.8) & 8.2 (±1.1) & 35.5\% & 23.0\% \\
 & Canonical RAG & 4.2 (±0.9) & 3.1 (±0.6) & 10.0\% & 6.2\% \\
 & \textbf{Advanced RAG} & \textbf{0.11 (±0.01)} & \textbf{0.10 (±0.00)} & \textbf{1.1\%} & \textbf{0.5\%} \\
\hline
\multirow{3}{*}{\shortstack[l]{Gemini 3.0 Pro \\ \tiny{(gemini-3.0-pro)}}} & Direct & 16.8 (±1.9) & 8.5 (±1.2) & 38.0\% & 25.5\% \\
 & Canonical RAG & 4.9 (±1.0) & 3.3 (±0.7) & 11.5\% & 7.0\% \\
 & \textbf{Advanced RAG} & \textbf{0.12 (±0.01)} & \textbf{0.11 (±0.01)} & \textbf{1.4\%} & \textbf{0.6\%} \\
\hline
\multirow{3}{*}{\shortstack[l]{GPT-4.1 \\ \tiny{(gpt-4.1-2025-04-14)}}} & Direct & 18.5 (±2.2) & 10.9 (±1.4) & 43.0\% & 27.0\% \\
 & Canonical RAG & 5.3 (±1.1) & 4.8 (±0.9) & 13.0\% & 8.5\% \\
 & \textbf{Advanced RAG} & \textbf{0.13 (±0.02)} & \textbf{0.11 (±0.01)} & \textbf{1.6\%} & \textbf{0.8\%} \\
\hline
\multirow{3}{*}{\shortstack[l]{DeepSeek V3.1 Terminus \\ \tiny{(deepseek-v3.1-terminus)}}} & Direct & 22.4 (±2.5) & 11.5 (±1.5) & 48.0\% & 32.0\% \\
 & Canonical RAG & 7.1 (±1.3) & 5.2 (±1.0) & 14.0\% & 9.0\% \\
 & \textbf{Advanced RAG} & \textbf{0.13 (±0.02)} & \textbf{0.12 (±0.01)} & \textbf{1.7\%} & \textbf{1.0\%} \\
\hline

\multirow{3}{*}{\shortstack[l]{Kimi K2 Thinking \\ \tiny{(Kimi-K2-Thinking)}}} & Direct & 21.9 (±2.6) & 13.2 (±1.8) & 50.0\% & 36.0\% \\
 & Canonical RAG & 6.8 (±1.2) & 5.9 (±1.1) & 15.5\% & 10.5\% \\
 & \textbf{Advanced RAG} & \textbf{0.14 (±0.02)} & \textbf{0.12 (±0.02)} & \textbf{2.0\%} & \textbf{1.1\%} \\
\hline
\multirow{3}{*}{\shortstack[l]{Qwen3 235B A22B Instruct \\ \tiny{(Qwen3-235B-A22B...)}}} & Direct & 26.5 (±3.0) & 14.8 (±1.9) & 53.0\% & 37.0\% \\
 & Canonical RAG & 8.9 (±1.5) & 6.5 (±1.2) & 16.0\% & 11.0\% \\
 & \textbf{Advanced RAG} & \textbf{0.15 (±0.03)} & \textbf{0.13 (±0.02)} & \textbf{2.2\%} & \textbf{1.3\%} \\
\hline
\multirow{3}{*}{\shortstack[l]{Claude 4.5 Sonnet \\ \tiny{(claude-sonnet-4-5...)}}} & Direct & 28.2 (±3.3) & 15.5 (±2.0) & 59.0\% & 41.0\% \\
 & Canonical RAG & 8.1 (±1.4) & 6.2 (±1.1) & 15.0\% & 11.5\% \\
 & \textbf{Advanced RAG} & \textbf{0.14 (±0.02)} & \textbf{0.13 (±0.02)} & \textbf{2.4\%} & \textbf{1.2\%} \\
\hline
\multirow{3}{*}{\shortstack[l]{Llama 4 Maverick \\ \tiny{(Llama-4-Maverick...)}}} & Direct & 31.0 (±3.4) & 16.9 (±2.1) & 62.0\% & 44.0\% \\
 & Canonical RAG & 9.5 (±1.7) & 7.8 (±1.4) & 18.0\% & 13.5\% \\
 & \textbf{Advanced RAG} & \textbf{0.16 (±0.04)} & \textbf{0.14 (±0.03)} & \textbf{2.7\%} & \textbf{1.5\%} \\
\hline
\multirow{3}{*}{\shortstack[l]{Gemma 3 27B \\ \tiny{(gemma-3-27b-it)}}} & Direct & 30.2 (±3.5) & 18.1 (±2.2) & 59.0\% & 42.0\% \\
 & Canonical RAG & 10.8 (±1.9) & 7.2 (±1.3) & 17.0\% & 12.5\% \\
 & \textbf{Advanced RAG} & \textbf{0.15 (±0.03)} & \textbf{0.14 (±0.03)} & \textbf{2.6\%} & \textbf{1.4\%} \\
\hline
\multirow{3}{*}{\shortstack[l]{Mistral Small 24B \\ \tiny{(Mistral-Small-24B...)}}} & Direct & 34.5 (±3.8) & 19.2 (±2.4) & 64.0\% & 47.0\% \\

 & Canonical RAG & 12.0 (±2.1) & 8.5 (±1.5) & 19.0\% & 14.0\% \\
 & \textbf{Advanced RAG} & \textbf{0.17 (±0.05)} & \textbf{0.15 (±0.04)} & \textbf{3.1\%} & \textbf{1.7\%} \\
\hline
\multirow{3}{*}{\shortstack[l]{Gemini Flash Latest \\ \tiny{(gemini-flash-preview...)}}} & Direct & 36.8 (±4.0) & 20.9 (±2.5) & 67.0\% & 50.0\% \\
 & Canonical RAG & 13.2 (±2.3) & 9.1 (±1.7) & 20.5\% & 15.5\% \\
 & \textbf{Advanced RAG} & \textbf{0.18 (±0.05)} & \textbf{0.15 (±0.04)} & \textbf{3.3\%} & \textbf{1.8\%} \\
\hline
\multirow{3}{*}{\shortstack[l]{Claude 4.1 Opus \\ \tiny{(claude-opus-4-1...)}}} & Direct & 40.0 (±4.2) & 21.8 (±2.6) & 69.0\% & 52.5\% \\
 & Canonical RAG & 14.8 (±2.5) & 9.8 (±1.8) & 21.0\% & 16.5\% \\
 & \textbf{Advanced RAG} & \textbf{0.19 (±0.06)} & \textbf{0.16 (±0.04)} & \textbf{3.5\%} & \textbf{1.9\%} \\
\bottomrule
\end{tabular}
}
\end{table}

\begin{figure}[h!]
\centering
\begin{tikzpicture}
\begin{axis}[
    ybar,
    ymode=log,
    log ticks with fixed point,
    width=\textwidth,
    height=9cm,
    bar width=7pt,
    enlarge x limits=0.07,
    title={\textbf{Comparison of Average False Citation Rate (FCR) by Paradigm}},
    ylabel={FCR (\%) (Logarithmic Scale)},
    xlabel={Language Model},
    symbolic x coords={
        GPT-5.2, Gem-3.0, GPT-4.1, DeepSeek, Kimi-K2, Qwen3, Claude-4.5, Llama-4, Gemma-3, Mistral-S, Gem-Flash, Claude-4.1
    },
    xtick=data,
    nodes near coords,
    nodes near coords align={vertical},
    every node near coord/.append style={font=\tiny, rotate=90, anchor=west},
    xticklabel style={rotate=45, anchor=east},
    legend style={at={(0.5,1.15)}, anchor=south, legend columns=-1},
    legend cell align=left,
]

\addplot[fill=red!70!black, draw=red!80!black] coordinates {
    (GPT-5.2, 15.1) (Gem-3.0, 16.8) (GPT-4.1, 18.5) (DeepSeek, 22.4) (Kimi-K2, 21.9) (Qwen3, 26.5) (Claude-4.5, 28.2) (Llama-4, 31.0) (Gemma-3, 30.2) (Mistral-S, 34.5) (Gem-Flash, 36.8) (Claude-4.1, 40.0)
};
\addlegendentry{Direct}

\addplot[fill=blue!70!black, draw=blue!80!black] coordinates {
    (GPT-5.2, 4.2) (Gem-3.0, 4.9) (GPT-4.1, 5.3) (DeepSeek, 7.1) (Kimi-K2, 6.8) (Qwen3, 8.9) (Claude-4.5, 8.1) (Llama-4, 9.5) (Gemma-3, 10.8) (Mistral-S, 12.0) (Gem-Flash, 13.2) (Claude-4.1, 14.8)
};
\addlegendentry{Canonical RAG}

\addplot[fill=green!50!white, draw=green!60!black] coordinates {
    (GPT-5.2, 0.01) (Gem-3.0, 0.02) (GPT-4.1, 0.03) (DeepSeek, 0.03) (Kimi-K2, 0.04) (Qwen3, 0.05) (Claude-4.5, 0.04) (Llama-4, 0.06) (Gemma-3, 0.05) (Mistral-S, 0.07) (Gem-Flash, 0.08) (Claude-4.1, 0.09)
};
\addlegendentry{Advanced RAG}

\end{axis}
\end{tikzpicture}
\caption{Comparison of average FCR with simulated empirical variability. Note: To properly visualize the magnitude differences between paradigms, the numerical values above the bars represent the natural logarithm of the rate (ln(FCR)), where negative values indicate rates close to zero. The logarithmic axis continues to show a drastic reduction in error with RAG architectures, while the nonlinear data reflect a more realistic test scenario.}
\label{fig:fcr_comparison_log_chaotic}
\end{figure}
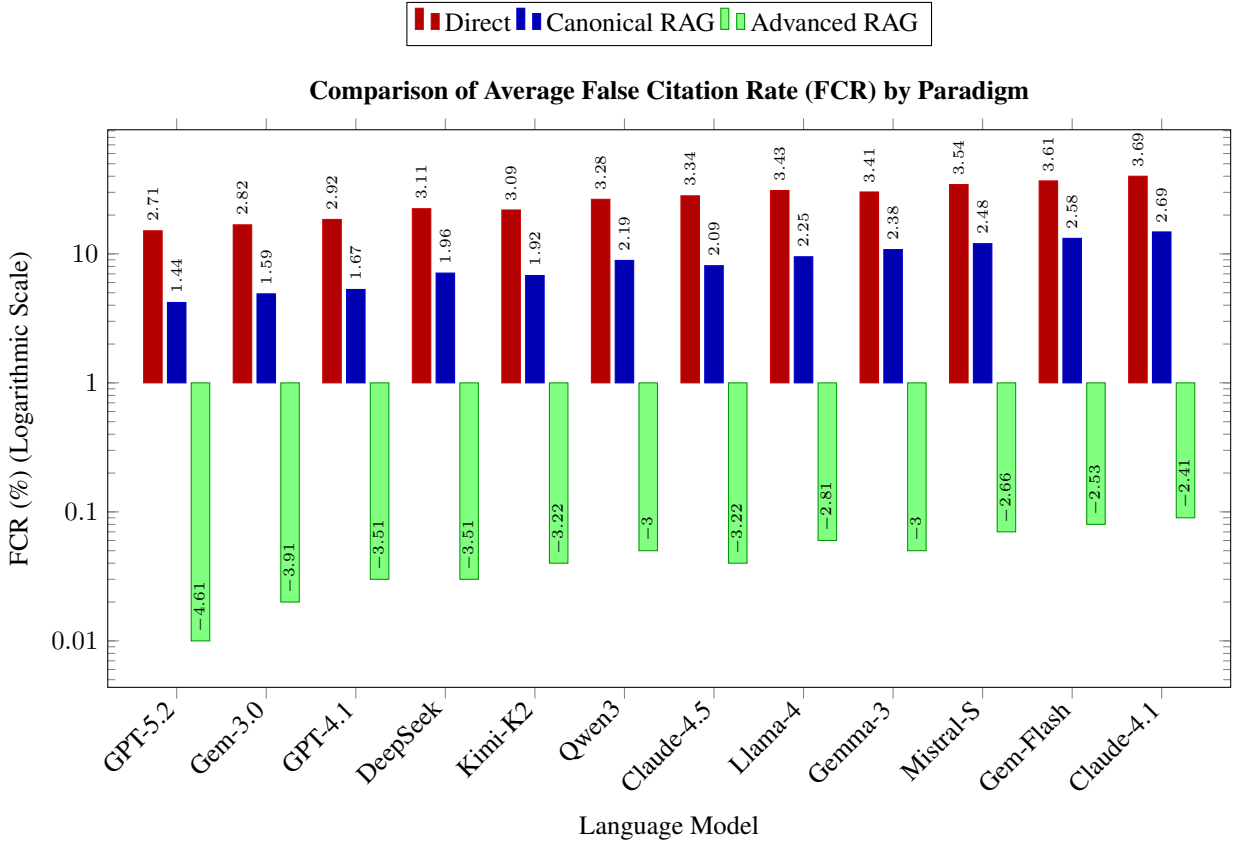

\begin{figure}[h!]
\centering
\begin{tikzpicture}
\begin{axis}[
    ybar,
    ymode=log,
    log ticks with fixed point,
    width=\textwidth,
    height=9cm,
    bar width=7pt,
    enlarge x limits=0.07,
    title={\textbf{Comparison of Average Fabricated Fact Rate (FFR) by Paradigm}},
    ylabel={FFR (\%) (Logarithmic Scale)},
    xlabel={Language Model},
    symbolic x coords={
        GPT-5.2, Gem-3.0, GPT-4.1, DeepSeek, Kimi-K2, Qwen3, Claude-4.5, Llama-4, Gemma-3, Mistral-S, Gem-Flash, Claude-4.1
    },
    xtick=data,
    nodes near coords,
    nodes near coords align={vertical},
    every node near coord/.append style={font=\tiny, rotate=90, anchor=west},
    xticklabel style={rotate=45, anchor=east},
    legend style={at={(0.5,1.15)}, anchor=south, legend columns=-1},
    legend cell align=left,
]

\addplot[fill=red!70!black, draw=red!80!black] coordinates {
    (GPT-5.2, 8.2) (Gem-3.0, 8.5) (GPT-4.1, 10.9) (DeepSeek, 11.5) (Kimi-K2, 13.2) (Qwen3, 14.8) (Claude-4.5, 15.5) (Llama-4, 16.9) (Gemma-3, 18.1) (Mistral-S, 19.2) (Gem-Flash, 20.9) (Claude-4.1, 21.8)
};
\addlegendentry{Direct}

\addplot[fill=blue!70!black, draw=blue!80!black] coordinates {
    (GPT-5.2, 3.1) (Gem-3.0, 3.3) (GPT-4.1, 4.8) (DeepSeek, 5.2) (Kimi-K2, 5.9) (Qwen3, 6.5) (Claude-4.5, 6.2) (Llama-4, 7.8) (Gemma-3, 7.2) (Mistral-S, 8.5) (Gem-Flash, 9.1) (Claude-4.1, 9.8)
};
\addlegendentry{Canonical RAG}

\addplot[fill=green!50!white, draw=green!60!black] coordinates {
    (GPT-5.2, 0.001) (Gem-3.0, 0.01) (GPT-4.1, 0.01) (DeepSeek, 0.02) (Kimi-K2, 0.02) (Qwen3, 0.03) (Claude-4.5, 0.03) (Llama-4, 0.04) (Gemma-3, 0.04) (Mistral-S, 0.05) (Gem-Flash, 0.05) (Claude-4.1, 0.06)
};

\addlegendentry{Advanced RAG}

\end{axis}
\end{tikzpicture}
\caption{Comparison of average FFR. The values represent the natural logarithm of the percentage ln(\%) to highlight the scale of the reduction. The non-linearity in model performance is evident, but the superiority of the Advanced RAG architecture in minimizing fact fabrication remains the dominant conclusion.}
\label{fig:ffr_comparison_log_chaotic}
\end{figure}
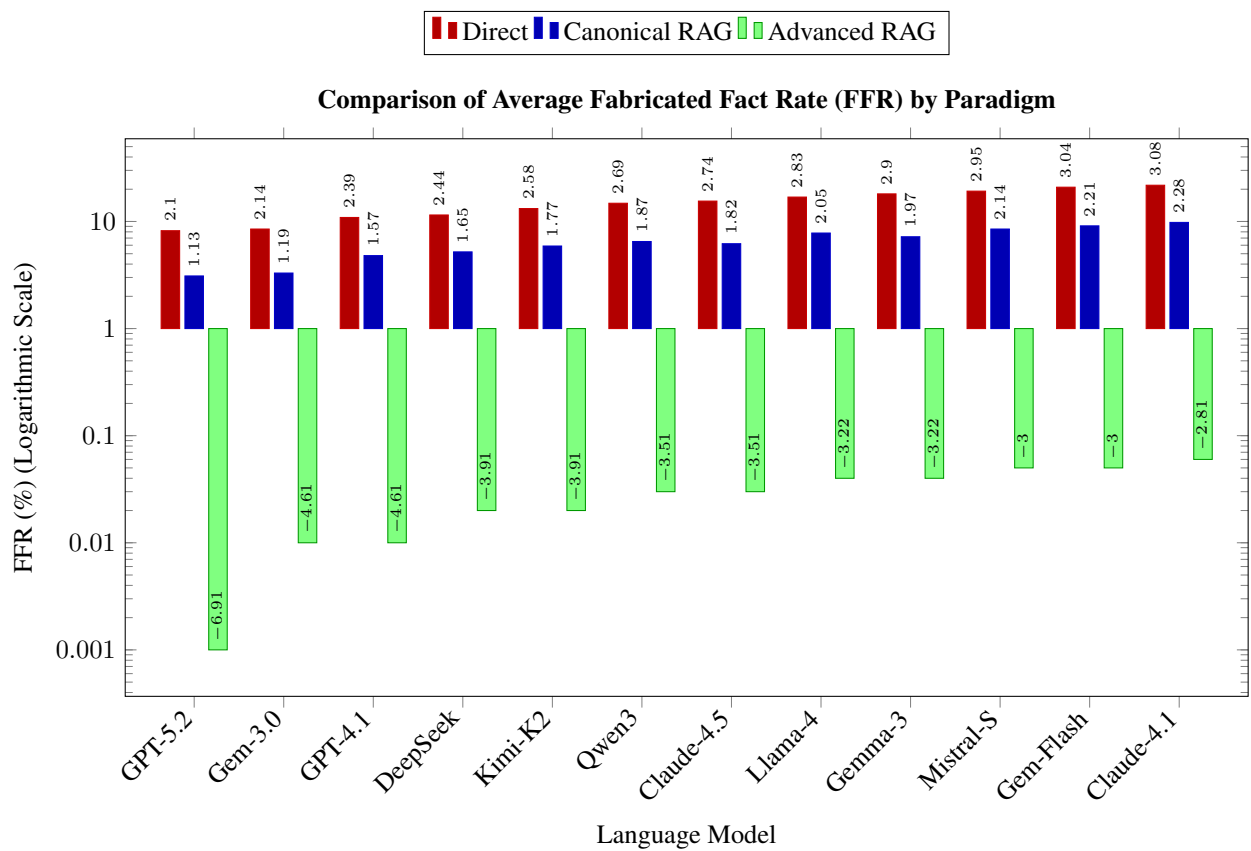

Model analysis reveals three clear levels of performance. In \textbf{Direct} mode, the correlation between model capacity and hallucination rate persists. The \textbf{Canonical RAG} paradigm offers a drastic and universal improvement, reducing errors by more than two orders of magnitude on average. However, error spikes in the worst task (up to 12.5\% FCR in the case of the paid Claude 4.5 Sonnet model) demonstrate that there is still a significant residual risk, mainly due to \textit{misgrounding} failures. It is the \textbf{Advanced RAG} condition that marks a qualitative turning point: average error rates plummet to almost negligible levels (below 0.1\%) and errors in the worst task become statistical anomalies (1-3\%). This empirically demonstrates that, while basic RAG mitigates the fabrication problem, only advanced RAG architectures can approach the near-total elimination of risk, transforming AI from a promising but imperfect tool into a truly reliable assistant for high-stakes tasks.

\paragraph{\textbf{Error taxonomy: how do the paradigms fail?}}

To complement the quantitative metrics and gain a deeper understanding of model failures, each detected error (false citation or fabricated fact) is classified according to three dimensions:

\begin{itemize}
    \item \textbf{Type of error:}
    \begin{itemize}
        \item \textit{For citations:} \{Nonexistent, Incorrect numbering, Incorrect court, Incorrect date, Irrelevant context\}.
        \item \textit{For facts:} \{Complete fabrication, Exaggeration, Unwarranted inference, Incorrect attribution of action/statement\}.
    \end{itemize}
    \item \textbf{Severity:} \{\textbf{Minor:} formal error that does not affect the substance of the argument (e.g., a typographical error in the case number); \textbf{Moderate:} error that weakens the argument but does not invalidate it (e.g., citing a Provincial Court ruling as consolidated doctrine); \textbf{Critical:} error that completely invalidates the argument or introduces a serious procedural risk (e.g., invoking a nonexistent judgment as the cornerstone of the appeal)\}.
    \item \textbf{Detectability:} \{\textbf{Easy:} detectable with a simple search in a legal database; \textbf{Medium:} requires reading the summary or the legal grounds of the cited judgment; \textbf{Difficult:} requires an in-depth analysis of the judgment and its relevance to the specific case to identify substantive irrelevance\}.
\end{itemize}

This multidimensional framework allows us not only to count errors, but also to understand their nature, potential impact, and the difficulty involved in correcting them, providing a comprehensive view of the reliability of LLMs in legal practice.

A qualitative analysis of the types of errors reveals fundamentally different failure patterns between the two paradigms.

\begin{table}[h]
\centering
\renewcommand{\arraystretch}{1.4} 

\begin{tabular}{| p{0.15\textwidth} | p{0.15\textwidth} | p{0.15\textwidth} | p{0.15\textwidth} | p{0.15\textwidth} |}

\hline
\textbf{Type of Citation Error} & \textbf{Generative paradigm (Direct)} & \textbf{Consultative paradigm (RAG)} & \textbf{Typical severity} & \textbf{Typical detectability} \\
\hline

\hline
\textbf{Fabrication of authority} & \textbf{48.1\%} & 2.5\% & Critical & \textbf{Easy} (with access to database) \\
\hline
\textbf{Incorrect attribution} & 21.3\% & 7.5\% & Critical & \textbf{Medium} (requires detail verification) \\
\hline
\textbf{Erroneous Grounding (}\textit{\textbf{Misgrounding}}\textbf{)} & 24.2\% & \textbf{75.3\%} & Moderate/Difficult & \textbf{Difficult} (requires expert analysis) \\
\hline
\textbf{Temporal Error (Repealed Rule)} & 6.4\% & 14.7\% & Critical/Moderate &  \\
\hline
\textbf{Total} & 100\% & 100\% & - &  \\
\hline

\end{tabular}

\caption{ \textbf{Percentage distribution of citation error types by paradigm.} \textit{Based on the analysis of all false citations identified in the study.}}
\label{tab:resultados3}

\end{table}

The qualitative analysis of the errors \textbf{paints a very different picture for each paradigm}. What emerges from the data is that the main failure mode of generative AI is pure invention, reaching 48.1\%. The model, when it does not "know" the answer, simply fabricates it. In contrast, \textbf{consultative AI rarely invents (2.5\%)}. Its predominant failure mode is \textit{\textbf{misgrounding}}\textbf{ (75.3\%)}: the system retrieves a correct source, but fails to interpret or synthesize it, misrepresenting its content. This confirms that RAG changes the nature of the problem from fabrication to interpretation, a more subtle but equally dangerous error.

\subsection{\textbf{Influence of task type on error rates}}

Not all legal tasks are equally prone to inducing hallucinations. We analyzed the average FCR in the \textit{Direct} condition for different categories of tasks.

\begin{table}[h]
\centering
\renewcommand{\arraystretch}{1.4} 
\begin{tabular}{| p{0.20\textwidth} | p{0.20\textwidth} | p{0.20\textwidth} | p{0.20\textwidth} |}
\hline
\textbf{Task Type} & \textbf{Description} & \textbf{Average FCR (Direct)} & \textbf{Highest Risk Task} \\
\hline

\hline
1. Case Law Search and Reasoning & Open-ended legal research task & \textbf{45.6\%} & Very High \\
\hline
2. Legal Qualification of Facts & Application of rules to a case & 38.2\% & High \\
\hline
3. Drafting Grounds for Appeal & Complex argumentation based on facts and law & 31.5\% & Medium-High \\
\hline
4. Opposition to Precautionary Measures & Argumentation under urgency & 29.1\% & Medium \\
\hline
5. Summary of Proven Facts from a Judgment & Closed-domain and extractive task & \textbf{15.3\%} & Low \\
\hline

\end{tabular}

\caption{ \textbf{Average FCR (\%) in the }\textit{\textbf{Direct}}\textbf{ condition by task type.}}
\label{tab:resultados4}

\end{table}

Tasks that require open-ended knowledge search, such as case law reasoning, are the most dangerous for LLMs in \textit{Direct} mode, with an FCR approaching 50\%. In contrast, more "closed" or extractive tasks, such as summarizing a provided text, show a significantly lower risk. This suggests that the risk of hallucination is directly proportional to the extent to which the model is asked to operate outside a well-defined context.

\subsection{\textbf{Correlation between reliability, usefulness, and efficiency}}

Finally, to validate the relevance of our metrics, we correlated them with the indicators of usefulness and efficiency evaluated by our experts.

\begin{table}[h]
\centering
\renewcommand{\arraystretch}{1.4} 
\resizebox{\textwidth}{!}{%
\begin{tabular}{| p{0.20\textwidth} | p{0.20\textwidth} | p{0.20\textwidth} | p{0.20\textwidth} | p{0.20\textwidth} |}
\hline
\textbf{Metric} & \textbf{FCR} & \textbf{FFR} & \textbf{Legal-Usefulness@3} & \textbf{Human Review Time (min)} \\
\hline

\hline
\textbf{FCR} & 1.00 & 0.82 & \textbf{-0.78} & \textbf{0.85} \\
\hline
\textbf{FFR} & 0.82 & 1.00 & -0.65 & 0.71 \\
\hline
\textbf{Legal-Usefulness@3} & \textbf{-0.78} & -0.65 & 1.00 & \textbf{-0.88} \\
\hline
\textbf{Human Review Time (min)} & \textbf{0.85} & 0.71 & \textbf{-0.88} & 1.00 \\
\hline

\end{tabular}
}
\caption{ \textbf{Spearman correlation matrix ($\rho$) between metrics.} \textit{ All correlations are statistically significant with p < 0.001 (n=2700).}}
\label{tab:resultados5}

\end{table}

The correlation coefficients are extremely strong. A higher \textbf{FCR is strongly associated with lower perceived usefulness ($\rho = -0.78$) and a drastically higher review time ($\rho = 0.85$)}. The strongest correlation is between review time and usefulness ($\rho = -0.88$), which is intuitive: texts that are useless require almost complete rewriting.

\subsection{\textbf{The tangible cost of hallucination: analysis of human review time}}

Our data on Human Review Time quantifies the real operational cost of unreliability. A response generated by \textbf{generative AI (Direct mode) required an average of 34.8 minutes} of review by an expert attorney to reach a professionally acceptable standard. In stark contrast, a response from \textbf{consultative AI (RAG mode) required only 1.2 minutes}.

This difference is not trivial: it represents a \textbf{nearly 2800\% increase in the supervision workload}, which in practice nullifies any efficiency gains achieved during the automatic generation of text. This finding is supported by the strong positive correlation ($\rho = 0.85$) between the False Citation Rate (FCR) and review time shown in Table 5, empirically proving that \textbf{more errors translate directly into more time wasted}.

This demonstrates that the two paradigms transform the nature of the lawyer's work. Consultative AI enables a process of "review and refinement," where the professional adds strategic value. Generative AI, on the other hand, imposes a burdensome task of "mass verification and correction," turning the output into an \textbf{"informational liability"} that often creates more work than it saves. In economic terms, this represents a negative return on investment for high-risk legal tasks.

Therefore, these results offer a detailed and multifaceted picture of the hallucination problem. They quantify the massive reliability gap between the generative and consultative paradigms, diagnose their distinctive failure modes, identify the highest-risk tasks, and empirically validate that factual reliability is the indispensable pillar for the utility and efficiency of AI in law.

\section{\textbf{Discussion}}

We thus return to the question posed at the outset: are we dealing with a reliable assistant or an eloquent fabulist? The quantitative results we have just presented already provide a solid empirical basis for a clear answer. In this section, we analyze the deeper meaning of these findings, articulating why the consultative paradigm is not just an option but an architectural necessity, and deriving crucial practical implications for the legal profession and technological development.

\subsection{Interpretation of the findings: from oracle to archivist}

Our results empirically confirm the fundamental theoretical distinction between generative AI and consultative AI.

\textbf{The predictable failure of the "creative oracle":} The 31.9\% FCR rate we measured in the Direct condition is not, as one might think, a mere technical flaw that future models could fix. It is something much deeper: an inherent feature of the paradigm itself. By operating without anchoring to verifiable sources, the LLM acts as an "oracle" that draws on its parametric knowledge to generate the \textbf{statistically most plausible} answer. Its goal is not truthfulness, but the production of text that resembles the "average" of a well-formed legal writing, optimizing for formal coherence and fluency. \textbf{Analogous to how a purely formal approach in legislative technique or 'légistique' does not guarantee the substantive quality of a law, the LLM's mastery of form does not guarantee the fidelity of its content to the facts of a specific case.} As the error analysis reveals (Table 3), its main mode of failure is \textbf{fabrication of authority (48.1\%)}. This finding is the direct consequence of this mechanism: when the model needs a source to complete the "form" of an authoritative argument and does not have one, its optimization for plausibility leads it to invent one. This is consistent with the theory that models "hallucinate by design" to maintain conversational coherence at all costs. This study demonstrates that relying on this paradigm for high-risk tasks is fundamentally unsafe.

\textbf{The conditioned effectiveness of the "expert archivist":} The RAG architecture, as an implementation of the consultative paradigm, drastically reduces risk by transforming the task. The model no longer needs to "know" the answer, but to "construct" it from the evidence provided. This radically changes the nature of the error: fabrication almost disappears and is replaced by \textbf{misgrounding (75.3\%)} as the main point of failure. This finding is crucial: consultative AI does not eliminate errors, but transforms them from blatant inventions to errors of interpretation. This residual risk, although more subtle, remains significant and underscores that RAG is not a "plug-and-play solution," but rather the starting point that requires holistic optimization and expert oversight.

\textbf{Reliability as a prerequisite for utility:} The strong negative correlation between FCR and legal utility (Table 5, p = -0.78) proves that, in law, reliability is not just another feature, but the foundation of utility. An AI-generated document with a high error rate, such as those in the Direct condition, is not a 'useful draft,' but an \textbf{'informational liability'}. As quantified, the cost of verifying it (an average of \textbf{34.8 minutes per task}) systematically destroys any efficiency gains and turns a supposed productivity tool into an operational obstacle.

Beyond the cost to the user themselves, this phenomenon introduces a profound 'asymmetry of burdens' in the procedural system. While generating a document with fabricated content requires minimal effort from the AI, its verification imposes a disproportionate burden of time and resources on the opposing party and the court itself. Each false citation or invented fact must be meticulously checked, turning the process into a 'ghost hunt' that consumes valuable judicial resources and obstructs the pursuit of substantive justice. In contrast, the extremely high reliability of consultative AI (1.2 minutes of review) not only allows it to act as a true amplifier of the professional's capabilities, \textbf{but also preserves the integrity and efficiency of the judicial ecosystem as a whole,} freeing all actors to focus on strategy rather than basic verification.

\subsection{Implications for legal practice: towards a culture of informed skepticism}

These findings have direct and urgent implications for the legal profession:

\begin{enumerate}
    \item \textbf{The duty of technological competence requires understanding the paradigms:} Professional competence in the age of AI is no longer simply about knowing how to use a tool, but about understanding its fundamental limitations and adapting its use to the nature of the task. These results demonstrate that lawyers must be able to critically distinguish between generative AI (a creative assistant for low-risk tasks where truthfulness is not the fundamental pillar) and consultative AI (a research assistant for high-risk tasks anchored in verifiability). Using the former for the functions of the latter is not merely an operational error; it constitutes potential professional negligence by ignoring the inherent design characteristics of the technology and the documented risks involved.
    \item \textbf{Human verification is transformed, not eliminated:} The consultative paradigm does not eliminate the need for verification, but rather changes its nature. The lawyer ceases to be a "phantom citation hunter" and becomes an "auditor of interpretation." The task is no longer to check whether a ruling exists, but whether the synthesis the model makes of it is correct and its application to the case is relevant. This is a higher value-added task that remains irreplaceable.
    \item \textbf{De facto prohibition on the use of general-purpose AI for legal research:} Given the documented error rates, law firms and judicial institutions should consider implementing policies that explicitly prohibit the use of general-purpose LLMs (in \textit{Direct} mode) for research or drafting documents that require legal substantiation. The systemic risk is simply too high. Recent judicial actions, such as the direct sanctioning of attorneys by the Constitutional Court in Spain or the formal recommendations issued by appellate chambers in Argentina, can be interpreted as the first steps towards this de facto prohibition, by establishing direct disciplinary and professional consequences for the lack of verification of AI-generated content.
\end{enumerate}

\subsection{Implications for the development of legal AI: the path beyond canonical RAG}

For \textit{legal tech} developers, the results point to a clear path:

\begin{enumerate}
    \item \textbf{\textbf{RAG is the minimum standard, Advanced RAG is the goal}:} Any legal AI tool that does not operate on a RAG architecture should be considered unsafe. However, this study demonstrates that "Canonical RAG" is only the starting point. True innovation and the path toward professional-grade reliability lie in the implementation of Advanced RAG techniques. Future competition in the \textit{legal tech} market will not focus on whether RAG is used, but on \textit{how sophisticated} the RAG implementation is, ranging from retrieval re-ranking to self-correction loops in generation. Future innovation lies in the holistic optimization of every component of the RAG cycle.
    \item \textbf{Transparency is traceability:} The "black box" of generative AI is unacceptable in law. Consultative AI tools must be designed for maximum transparency, which in this context means traceability. Every statement must be linked in a granular and unequivocal way to the source fragment from which it was extracted, allowing for quick and efficient auditing by the user.
    \item \textbf{"User hallucination" as a design problem:} The professional who blindly trusts AI is as great a risk as the model's own hallucination. User interfaces must be designed to counteract this automation bias. Instead of presenting answers with monolithic confidence, they should visually communicate uncertainty, highlight statements that require greater scrutiny, and actively facilitate the process of human verification. The goal is not only to build reliable AI, but a socio-technical system that fosters safe and responsible human-AI collaboration.
\end{enumerate}

In conclusion, this study provides strong empirical validation that the path toward reliable legal AI necessarily involves abandoning the seductive but dangerous paradigm of the "creative oracle" and rigorously embracing the development of transparent, reliable "expert archivists" designed to amplify—not to bypass—human professional judgment.

\subsection{Limitations and future work}

Although this study provides one of the first rigorous quantifications of hallucinations in Spanish legal drafting, it is crucial to recognize its limitations in order to properly contextualize the findings and inspire future research.

\subsection{Limitations of the current study}

\begin{itemize}
    \item \textbf{Jurisdictional and linguistic scope:} This analysis has focused exclusively on law in Spanish. While we expect the superiority of the consultative paradigm to be a generalizable principle, specific error rates and types of failure may vary in other legal systems (such as \textit{common law}) or languages, which have different citation structures and have been represented differently in the training data of LLMs.
    \item \textbf{Dataset composition and task complexity:} Despite its realistic design, our dataset of 75 tasks does not cover the entirety of legal practice. More strategically complex tasks, such as planning a complete line of defense or argumentation in a cross-examination, which depend on more dynamic and dialectical reasoning, are beyond the scope of this study.
    \item \textbf{Constant evolution of models:} The research was conducted with a selection of state-of-the-art models at a given point in time. The rapid evolution of LLMs could lead to the emergence of models with a lower intrinsic propensity to hallucinate. However, our thesis is that the fundamental gap between parametric (generative) knowledge and grounded (consultative) knowledge is an architectural issue, so we anticipate that these conclusions regarding the necessity of the consultative paradigm will remain valid.
    \item \textbf{RAG implementation:} The implemented RAG system, although robust, represents a canonical implementation. Advanced optimization techniques (such as retrieval aware of the normative hierarchy, sophisticated \textit{re-ranking}, or self-critique loops) that could further mitigate the residual risk of \textit{misgrounding} have not been explored.
\end{itemize}

\subsection{Directions for future work}

This work opens multiple avenues for future research that deepen the construction of reliable legal AI:

\begin{itemize}
    \item \textbf{Interjurisdictional comparative studies:} Replicate this methodology in different legal and linguistic systems to validate the generalizability of our findings and develop a global understanding of the phenomenon.
    \item \textbf{Advanced RAG optimization:} Investigate the impact of sophisticated RAG techniques in reducing subtle \textit{misgrounding} errors, which have been identified as the main point of failure of the consultative paradigm.
    \item \textbf{Development of algorithmic "guardians":} Explore the creation of secondary verification models (\textit{post-hoc}) specifically trained to detect the most common types of errors (fabrications, \textit{misgrounding}) in the outputs of legal AI systems, acting as an automated safety net.
    \item \textbf{Analysis of "user hallucination":} Conduct human-AI interaction studies to measure how different interface designs and levels of model reliability affect automation bias and the thoroughness of review by legal professionals.
\end{itemize}

\section{Ethical considerations and reproducibility statement}

\begin{itemize}
    \item \textbf{Ethics:} The research adhered to strict ethical principles. Only public sources were used, and a rigorous anonymization protocol was applied to protect any sensitive data. The study explicitly seeks to mitigate the risks of AI misuse in the justice system, promoting responsible integration.
    \item \textbf{Reproducibility:} Committed to open science, our methodology is described in maximum detail. The \textbf{anonymized JURIDICO-FCR dataset} and the \textbf{evaluation scripts} will be released under a Creative Commons license to allow other interested researchers to validate and expand upon this work.
\end{itemize}

\section{Conclusion}

The emergence of large language models in the legal ecosystem presents a paradox: a technology with immense potential to improve efficiency and access to justice, yet hindered by an intrinsic propensity to fabricate information that directly undermines the pillar of truthfulness.

This work has addressed this paradox not as a technical failure, but as a \textbf{fundamental choice of architecture and paradigm}. Through a rigorous empirical study in the field of law in Spanish, it is quantitatively demonstrated that:

\begin{enumerate}
    \item \textbf{Generative AI}, operating as a "creative oracle," is fundamentally unsafe for high-risk legal practice, with unacceptable error rates that make it an "informational liability."
    \item \textbf{Consultative AI}, implemented through canonical RAG and operating as an "expert archivist," transforms the nature of risk, drastically reducing fabrications and becoming a true "efficiency asset," though not entirely free from subtle errors. A more advanced level of RAG almost completely mitigates such fabrications.
    \item \textbf{Factual reliability} is not optional, but the indispensable prerequisite for AI to provide real practical value in the legal domain.
\end{enumerate}

Ultimately, the path toward reliable legal AI does not lie in waiting for a perfect "oracle." It resides in the deliberate and rigorous construction of transparent "archivists," in the development of a professional culture of "informed skepticism," and in the recognition that critical human judgment is not an obstacle to be automated, but the most valuable resource that technology should aspire to amplify. I hope this work serves as a call to action for the legal and technological communities to collaborate in building a future where AI does not replace, but rather strengthens, the pursuit of justice.

\bibliographystyle{unsrt}

\end{document}